\documentclass[runningheads]{llncs}
\usepackage{graphicx}

\usepackage{tikz}
\usepackage{comment}
\usepackage{amsmath,amssymb} 
\usepackage{color}

\usepackage[accsupp]{axessibility}  

\usepackage[width=122mm,left=12mm,paperwidth=146mm,height=193mm,top=12mm,paperheight=217mm]{geometry}

\usepackage{multirow}
\usepackage{enumitem}
\usepackage{booktabs}

\usepackage{array}
\newcommand{\PreserveBackslash}[1]{\let\temp=\\#1\let\\=\temp}
\newcolumntype{C}[1]{>{\PreserveBackslash\centering}p{#1}}
\newcolumntype{R}[1]{>{\PreserveBackslash\raggedleft}p{#1}}
\newcolumntype{L}[1]{>{\PreserveBackslash\raggedright}p{#1}}

\begin{document}
	\pagestyle{headings}
	\mainmatter
	\def\ECCVSubNumber{6359}  
	
	\title{Transformer Scale Gate for Semantic Segmentation} 

	\titlerunning{Transformer Scale Gate for Semantic Segmentation}
	%
	\author{Hengcan Shi \and
	Munawar Hayat \and
	Jianfei Cai}
	\authorrunning{H. Shi et al.}
	%
	\institute{Monash University, Australia \\
	\email{\{hengcan.shi, munawar.hayat, jianfei.cai\}@monash.edu}}
	\maketitle
	
	\begin{abstract}
		
		Effectively encoding multi-scale contextual information is crucial for accurate semantic segmentation.
		Existing transformer-based segmentation models combine features across scales without any selection, where features on sub-optimal scales may degrade segmentation outcomes. Leveraging from the inherent properties of Vision Transformers, we propose a simple yet effective module, Transformer Scale Gate (TSG), to optimally combine multi-scale features.
		TSG exploits cues in self and cross attentions in Vision Transformers for the scale selection. TSG is a highly flexible plug-and-play module, and can easily be incorporated with any encoder-decoder-based hierarchical vision Transformer architecture. 
		Extensive experiments on the Pascal Context and ADE20K datasets demonstrate that our feature selection strategy achieves consistent gains.
		
	\end{abstract}

	\section{Introduction}
	Semantic segmentation aims to segment all objects including `things' and 'stuff' in an image and determine their categories. It is a challenging task in computer vision, and serves as a foundation for many higher-level tasks, such as scene understanding \cite{liang2016G-LSTM,shi2019scene}, object tracking \cite{muller2018trackingnet} and vision+language \cite{bai2018deep,shi2018key}. In recent years, Vision Transformers based on encoder-decoder architectures have been a new paradigm for semantic segmentation. The encoder consists of a series of multi-head self-attention modules to capture features of image patches, while the decoder has both self- and cross-attention modules to generate segmentation masks. Earlier works \cite{zheng2021rethinking,ranftl2021vision,strudel2021segmenter} usually use Vision Transformers designed for image classification to tackle the semantic segmentation problem, which only encode single-scale features. However, different from image classification that only recognizes one object in an image, semantic segmentation is generally expected to extract numerous objects of different sizes. It is hard to segment and recognize these various objects by only single-scale features.
	
	\begin{figure}[h]
		\centerline{\includegraphics[scale=0.25]{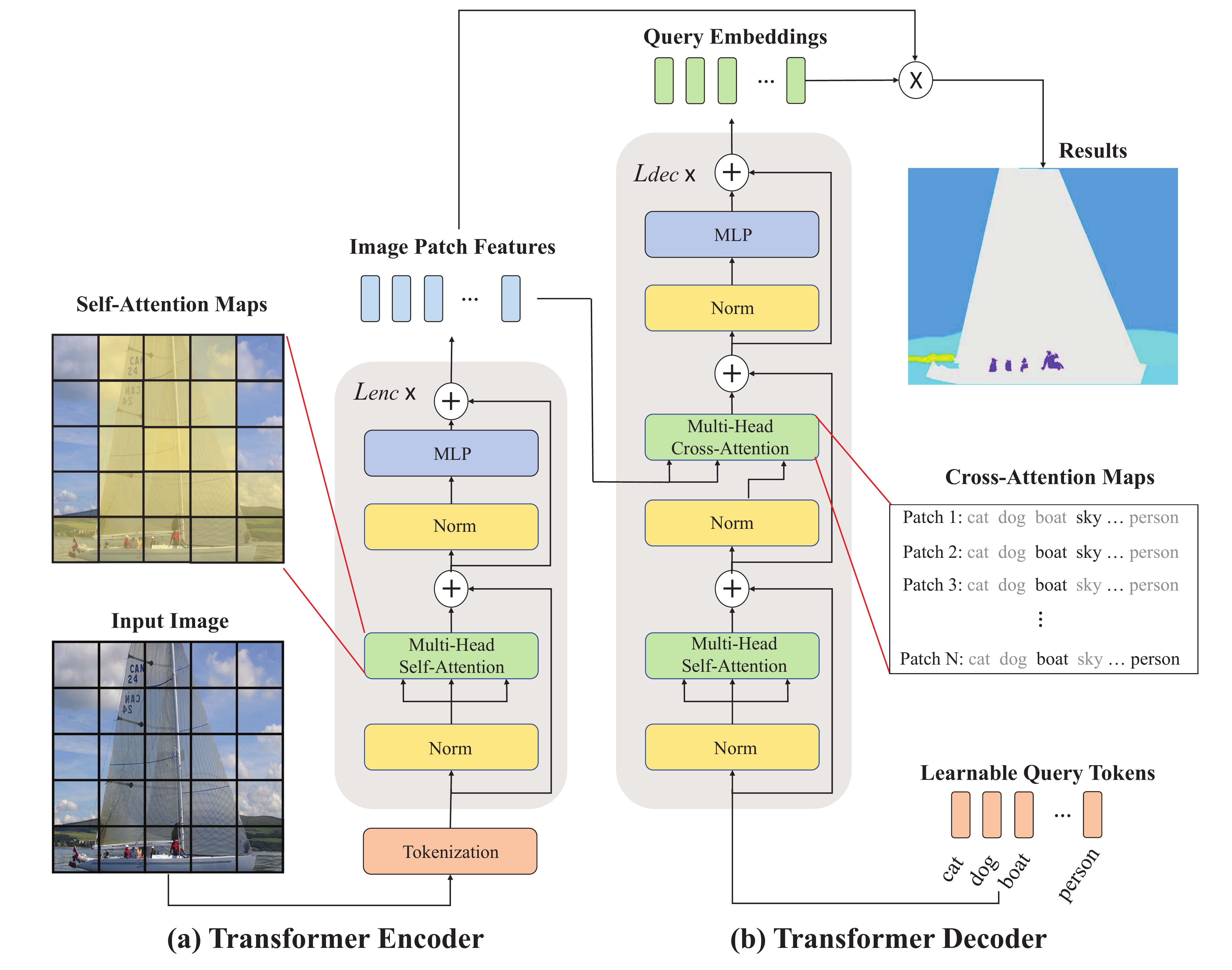}}
		\caption{Illustration of Vision Transformers for semantic segmentation. (a) The image is divided into multiple patches and input into the encoder. The encoder contains $L_{enc}$ blocks and outputs features for every image patch. (b) The decoder takes learnable query tokens as inputs, where each query is corresponding to an object category. The decoder with $L_{dec}$ blocks outputs query embeddings. Finally, segmentation results are generated by multiplying the image patch features and the query embeddings.}
		\label{fig_method1}
	\end{figure}
	
	Some recent methods \cite{liu2021swin,wang2021pyramid,wu2021p2t,gu2021multi,bousselham2021efficient} attempt to leverage multi-scale features to solve this problem. They first use hierarchical transformers such as Swin Transformer \cite{liu2021swin} and PVT \cite{wang2021pyramid} to extract multi-scale image features, and then combine them, e.g., by the pyramid pooling module (PPM) \cite{zhao2017pyramid} or the seminal feature pyramid network (FPN) \cite{lin2017feature} borrowed from CNNs. We argue that such feature combinations cannot effectively select an appropriate scale for each image patch. Features on sub-optimal scales may decrease the segmentation accuracy, which has been proved in CNN-based works \cite{DBLP:conf/cvpr/ChenYWXY16,shi2018boosting}.
	To address this issue, CNN-based methods \cite{DBLP:conf/cvpr/ChenYWXY16,shi2018boosting} design learnable models to select the optimal scales. Nevertheless, these models are complex, either use multiple networks \cite{DBLP:conf/cvpr/ChenYWXY16} or require scale labels \cite{shi2018boosting}, which decrease the network efficiency and may cause over-fitting.
	
	In this paper, we exploit inherent characteristics of Vision Transformers to guide the feature selection process. Specifically, our design is inspired from the following observations: \textbf{(1)} As shown in Fig.~\ref{fig_method1} (a), the self-attention module in the transformer encoder learns the correlations among image patches. In principle, if an image patch is correlated to a mass of patches, small-scale (low-resolution) features should be preferred for segmenting this patch, because small-scale features have large effective receptive fields \cite{xie2021segformer}, and vice versa.
	\textbf{(2)} In the transformer decoder, the cross-attention module models correlations between patch-query pairs, as shown in Fig.~\ref{fig_method1} (b), where the queries 
	are object categories. If an image patch is correlated to multiple queries, it indicates that this patch contains multiple objects and needs large-scale (high-resolution) features for fine-grained segmentation. On the contrary, an image patch correlated to only a few queries needs small-scale (low-resolution) features to avoid over-segmentations.
	\textbf{(3)} The above observations can guide design choices for not only category-query-based decoders but also object-query-based decoders. In object-query-based decoders, each query token corresponds to an object instance. Therefore, the cross-attention module extracts relationships between patch-object pairs. Many high cross-attention values also indicate the image patch contains multiple objects and needs high-resolution features.
	
	From these observations and analyses, we propose a novel Transformer Scale Gate (TSG) module that takes correlation maps in self- and cross-attention modules as inputs, and predicts weights of multi-scale features for each image patch. Our TSG is a simple design only with a few lightweight linear layers.
	We further extend TSG to a TSGE module and a TSGD module, which leverage our TSG weights to optimize multi-scale features in transformer encoders and decoders, respectively. 
	TSGE employs a pyramid structure to refine multi-scale features in the encoder by the self-attention guidance, while TSGD fuses these features in the decoder based on the cross-attention guidance. Experimental results on two datasets, Pascal Context \cite{mottaghi2014role} and ADE20K \cite{zhou2017scene}, show that the proposed modules consistently achieve gains, up to 4.3\% and 3.1\% in terms of mIoU, compared with Swin Transformer based baseline~\cite{liu2021swin}.
	
	Our main contributions can be summarized as follows: \textbf{(1)} To the best of our knowledge, this is the first work to exploit inner properties of Vision Transformers for multi-scale feature selection. We analyze the properties of Vision Transformers and design TSG for the selection. \textbf{(2)} We propose TSGE and TSGD in the encoder and decoder in Vision Transformers, respectively, which leverage our TSG to improve the semantic segmentation performance. \textbf{(3)} Our extensive experiments and ablations show that the proposed modules obtain significant improvements on two semantic segmentation datasets.
	
	\section{Related Work}
	\textbf{CNN-based semantic segmentation methods} \cite{long2015fully,shi2018hierarchical,noh2015learning} usually use the fully convolutional network (FCN) \cite{long2015fully}, which formulates the semantic segmentation task as a pixel-wise classification problem and designs an encoder-decoder structure. The encoder extracts features of each pixel in the image, while the decoder labels every pixel. 
	Noh \emph{et al.} \cite{noh2015learning} design a deconvolutional decoder to gradually restore more details, which is the mirror of the CNN encoder. Although these approaches make significant progress for recognizing objects of fixed size ranges, they struggle to segment objects of diverse sizes. To segment variable sized objects, prior works \cite{DBLP:conf/cvpr/LinMSR17,zhao2017pyramid,chen2018encoder,lin2017feature,badrinarayanan2017segnet,xiao2018unified,shi2018boosting,DBLP:conf/cvpr/ChenYWXY16} propose to use multi-scale strategies. Lin \emph{et al.} \cite{DBLP:conf/cvpr/LinMSR17} directly resize the input image into multiple scales and generate multiple predictions, which are then assembled together. Zhao \emph{et al.} \cite{zhao2017pyramid}, Chen \emph{et al.} \cite{chen2018encoder} and Lin \emph{et al.} \cite{lin2017feature} extract multi-scale features in the encoder by pyramid pooling module (PPM), pyramid atrous convolutions and feature pyramid network (FPN), respectively, and then combine multi-scale features to predict segmentation results. Xiao \emph{et al.} \cite{xiao2018unified} use both PPM and FPN to capture multi-scale features. Some methods \cite{badrinarayanan2017segnet,ronneberger2015u} adopt the deconvolutional decoder \cite{noh2015learning} and incorporate multi-scale features into the decoder. All these methods only simply combine multi-scale features without any selection. While some existing works \cite{DBLP:conf/cvpr/ChenYWXY16,shi2018boosting} propose learnable modules to select multiple scales, they either require complex networks \cite{DBLP:conf/cvpr/ChenYWXY16} or scale labels \cite{shi2018boosting}, and are therefore not ideally suited for efficient scale selection.
	
	\noindent \textbf{Vision Transformers} have recently attracted increasing research interest and have become a new paradigm for semantic segmentation, thanks to their ability of modeling long-range dependencies. Ranftl \emph{et al.} \cite{ranftl2021vision} and Zheng \emph{et al.} \cite{zheng2021rethinking} employ ViT \cite{dosovitskiy2020image} as the encoder to extract single-scale features and use CNN-based decoders for semantic segmentation. Strudel \emph{et al.} \cite{strudel2021segmenter} design a transformer-based decoder, which takes categories as queries. Cheng \emph{et al.} \cite{cheng2021per} propose an object-query-based transformer decoder and combine it with a pixel-level decoder to predict segmentation results. These methods ignore the diversity of object sizes. Other recent works \cite{liu2021swin,dong2021cswin,yang2021focal,wang2021pyramid,wu2021p2t,gu2021multi,xie2021segformer} design pyramid architectures to obtain multi-scale features in the encoder and fuse them in the decoder for segmentation. Most of them \cite{liu2021swin,dong2021cswin,yang2021focal,wang2021pyramid,wu2021p2t} adopt PPM \cite{zhao2017pyramid} and/or FPN \cite{lin2017feature} for multi-scale feature fusion, while Xie \emph{et al.} \cite{xie2021segformer} and Gu \emph{et al.} \cite{gu2021multi} use lightweight concatenations.
	Instead of combining multi-scale features, Bousselham \emph{et al.} \cite{bousselham2021efficient} leverage multiple transformer decoders to generate multi-scale segmentation results and then aggregate these results. 
	However, these combinations cannot effectively select multi-scale features for each image patch. Different from these works, our method leverages intrinsic properties of Vision Transformers for multi-scale feature selection, and thus improves the semantic segmentation performance.

	\section{Proposed Method}
	
	\subsection{Vision Transformer for Semantic Segmentation}
	
	Vision Transformers typically contain an encoder and a decoder, as shown in Fig.~\ref{fig_method1}. An image is firstly split into multiple patches and every patch is embedded into a token.
	Let $\mathbf{Z}=\{\mathbf{z_{1}}, \mathbf{z_{2}},..., \mathbf{z_{N}}\}$ represent the set of tokens, where $N$ is the number of patches, $\mathbf{z_{n}} \in \mathbb{R}^{d_{Z}}~(n = 1,...,N)$ is the token vector of the $n$-th patch, and $d_{Z}$ is the dimension of token vectors. 
	
	\textbf{Encoder.} The encoder takes these tokens as inputs, and outputs the feature vector of every image patch. The key component in the encoder is the multi-head self-attention, which learns the long-range dependencies of image patches as follows:
	\begin{equation}
		\mathbf{Q_{i}} = Linear(\mathbf{Z}), ~~ \mathbf{K_{i}} = Linear(\mathbf{Z}), ~~ \mathbf{V_{i}} = Linear(\mathbf{Z})
	\end{equation}
	\begin{equation}
		\mathbf{A_{i}^{self}} = Softmax(\frac{\mathbf{Q_{i}}\mathbf{K_{i}}^{T}}{\sqrt{d_s}}) 
	\end{equation}
	\begin{equation}
		\mathbf{H_{i}} = \mathbf{A_{i}^{self}}\mathbf{V_{i}} 
	\end{equation}
	\begin{equation}
		\mathbf{O} = Linear(Concat(\mathbf{H_{1}}, ..., \mathbf{H_{h_{self}}}))
	\end{equation}
	where $\mathbf{Z} \in \mathbb{R}^{N \times d_{Z}}$ is the set of image tokens, $i=1,...,h_{self}$ 
	and $h_{self}$ is the number of heads in the multi-head self-attention module. $Linear(\cdot)$ means the linear layer; $Softmax(\cdot)$ is the Softmax function; and $Concat(\cdot)$ represents the concatenation. 
	$\mathbf{Q_{i}}, \mathbf{K_{i}}, \mathbf{V_{i}} \in \mathbb{R}^{N \times d_s}$ denote the query, key and value, respectively, where $d_s$ is their dimension. $\mathbf{A_{i}^{self}} \in \mathbb{R}^{N \times N}$ is the self-attention map, which models long-range dependencies between every patch pair in the image. $\mathbf{H_{i}} \in \mathbb{R}^{N \times d_s}$ is the feature map generated by the $i$-th attention head. Feature maps in all heads are concatenated into a single map and transformed by a linear layer to output the feature map $\mathbf{O} \in \mathbb{R}^{N \times d_{O}}$, where $d_{O}$ is its dimension.
	
	An encoder block composes of a multi-head self-attention module, a multilayer perceptron (MLP) and normalizations, as shown in Fig.~\ref{fig_method1} (a). By cascading $L_{enc}$ encoder blocks, we can obtain the encoder-output image patch feature map $\mathbf{F} \in \mathbb{R}^{N \times d_{F}}$, and $\mathbf{F}=\{\mathbf{f_{1}}, \mathbf{f_{2}},..., \mathbf{f_{N}}\}$, where $\mathbf{f_{n}} \in \mathbb{R}^{d_{F}}$ 
	is the encoder-output feature vector for the $n$-th image patch, and $d_{F}$ is the feature dimension. 
	
	\textbf{Our baseline decoder.} The inputs of the transformer decoder are a series of query tokens. Inspired by \cite{strudel2021segmenter}, we use $C$ query tokens $\mathbf{X}=\{\mathbf{x_{1}}, \mathbf{x_{2}},..., \mathbf{x_{C}}\}$, where $C$ is the number of object classes and each token $\mathbf{x_{c}} \in \mathbb{R}^{d_{F}} ~ (c=1,...,C)$ corresponds to a class. The dimension of query tokens is the same as image patch features. Different from \cite{strudel2021segmenter}, which also uses image patch features $\mathbf{F}$ as query tokens in the decoder, we take image patch features $\mathbf{F}$ as keys and values to reduce computational costs.
	
	As shown in Fig.~\ref{fig_method1} (b), the decoder includes multi-head self-attention modules and cross-attention modules. Similar to the encoder, self-attention modules take query tokens as inputs and learn their relationships. In contrast, cross-attention modules aim to capture relationships between each image patch and query token:
	\begin{equation}
		\mathbf{Q_{j}} = Linear(\mathbf{X}), \mathbf{K_{j}} = Linear(\mathbf{F}), \mathbf{V_{j}} = Linear(\mathbf{F})
	\end{equation}
	\begin{equation}
		\mathbf{A_{j}^{cross}} = Softmax(\frac{\mathbf{Q_{j}}\mathbf{K_{j}}^{T}}{\sqrt{d_c}}) 
	\end{equation}
	\begin{equation}
		\mathbf{H_{j}} = \mathbf{A_{j}^{cross}}\mathbf{V_{j}}  
	\end{equation}
	where $\mathbf{F} \in \mathbb{R}^{N \times d_{F}}$ represents the image patch feature matrix, $\mathbf{X} \in \mathbb{R}^{C \times d_{F}}$ is the query token matrix, $j=1,...,h_{cross}$ and $h_{cross}$ denotes the number of heads in the cross-attention module. Similar to self-attention modules, $\mathbf{Q_{j}}, \mathbf{K_{j}}, \mathbf{V_{j}} \in \mathbb{R}^{N \times d_c}$ are the $d_c$-dimension query, key and value, respectively. $\mathbf{A_{j}^{cross}} \in \mathbb{R}^{C \times N}$ captures relationships between every patch-class pair. $\mathbf{H_{j}} \in \mathbb{R}^{C \times d_c}$ represents the feature map in the $j$-th head, and features in all heads can be combined by Eq.~(4).

	Through $L_{dec}$ decoder blocks, our decoder outputs query embeddings $\mathbf{Y}=\{\mathbf{y_{1}}, \mathbf{y_{2}},..., \mathbf{y_{C}}\}$, where $\mathbf{Y} \in \mathbb{R}^{C \times d_{F}}$ and $\mathbf{y_{c}} \in \mathbb{R}^{d_{F}}$ 
	is the embedding vector of the $c$-th class. The segmentation result can be predicted by the matrix product of patch feature matrix $\mathbf{F}$ and class query embedding matrix $\mathbf{Y}$:
	\begin{equation}
		\mathbf{P} = Softmax(\frac{\mathbf{F}\mathbf{Y}^{T}}{\sqrt{d_{F}}}) 
	\end{equation}
	where $\mathbf{P} \in \mathbb{R}^{N \times C}$ contains the classification scores for every image patch. 
	The segmentation result is composed of these patch-wise classification results.

	\subsection{Transformer Scale Gate (TSG)}
	
	\begin{figure}[t]
		\centerline{\includegraphics[scale=0.33]{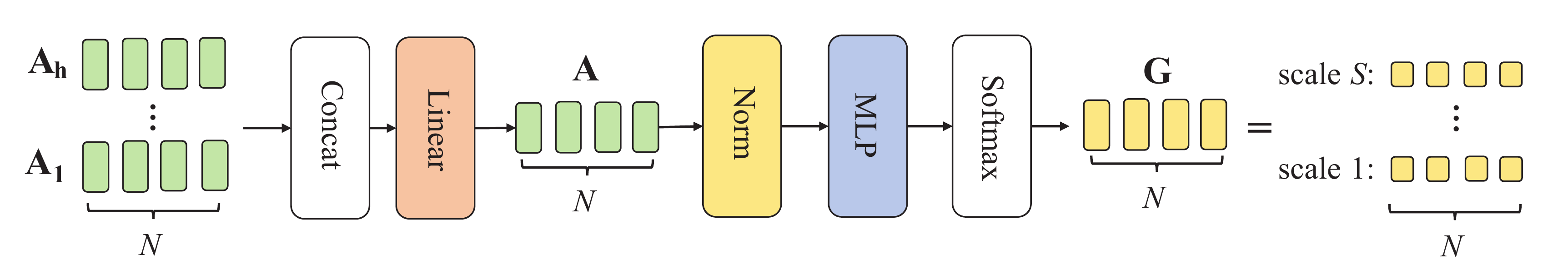}}
		\caption{Illustration of our Transformer Scale Gate (TSG).}
		\label{fig_method0}
	\end{figure}
	
	As illustrated in Fig.~\ref{fig_method2}, current Vision Transformer backbones \cite{liu2021swin,wang2021pyramid} use hierarchial structures to generate multi-scale features. We propose Transformer Scale Gate (TSG) to select features at the suitable scale for each image patch. 
	From Eq.~(2)\&(6), we observed that the self-attention map $\mathbf{A_{i}^{self}}$ reflects the correlation between an image patch and other patches, while the cross-attention map $\mathbf{A_{j}^{cross}}$ reflects the correlation between every image patch and object category. In $\mathbf{A_{i}^{self}}$, if an image patch is highly related to a large number of patches, it requires small-scale features (large effective receptive fields), and vice versa. In $\mathbf{A_{j}^{cross}}$, if an image patch is correlated to many object classes, this patch contains multiple objects and requires large-scale (high-resolution) features.
	
	Therefore, our TSG takes attention maps as inputs and generates gates for every scale. As shown in Fig.~\ref{fig_method0}, we first integrate multi-head attention maps into a single map $\mathbf{A} \in \mathbb{R}^{N \times d_{A}}$, where $d_{A}$ is its dimension. For self-attention modules in the encoder, attention maps are integrated as 
	\begin{equation}
		\mathbf{A} = Linear(Concat(\mathbf{A_{1}^{self}}, ..., \mathbf{A_{h_{self}}^{self}})).
	\end{equation}
	We concatenate attention maps in all heads and use a linear layer to project them to $d_{A}$ dimensions. 
	
	For cross-attention modules in the decoder, we first change the softmax in $\mathbf{A_{j}^{cross}}$. The original softmax is applied on the patch dimension (i.e., over $N$ image patches), which reflects the importance of patches for each object class. We change this softmax to the class dimension. The changed softmax reveals the importance of classes for each image patch, which is more suitable for our TSG. Let $\mathbf{\widetilde{A}_{j}^{cross}}$ represent the cross-attention map with the modified softmax. We concatenate the transposes of $\mathbf{\widetilde{A}_{j}^{cross}}$ in all heads and use a linear layer for the dimension transform, i.e.:
	\begin{equation}
		\mathbf{A} = Linear(Concat((\mathbf{\widetilde{A}_{1}^{cross}})^{T}, ..., (\mathbf{\widetilde{A}_{h_{cross}}^{cross}})^{T})).
	\end{equation}
	
	After the integration, our TSG generates multi-scale feature gates as
	\begin{equation}
		\mathbf{\widetilde{G}} = MLP(Norm(\mathbf{A})),
	\end{equation}
	\begin{equation}
		\mathbf{G} = Softmax(\mathbf{\widetilde{G}}).
	\end{equation}
	We employ a layer normalization to normalize $\mathbf{A}$, and use an MLP to predict the scale gates $\mathbf{\widetilde{G}} \in \mathbb{R}^{N \times S}$, where $S$ is the number of scales. Here, we use a two-layer MLP with GELU activation function.
	$\mathbf{\widetilde{G}}$ is normalized into $\mathbf{G}$ by a softmax function on the scale dimension. In matrix $\mathbf{G}$, value $g_{n,s}$ indicates the gate of the $s$-th scale for the $n$-th image patch. We next introduce how to use our scale gates to select features in the encoder and decoder.
	
	\subsection{Transformer Scale Gate in Encoder (TSGE)}
	
	\begin{figure}[t]
		\centerline{\includegraphics[scale=0.27]{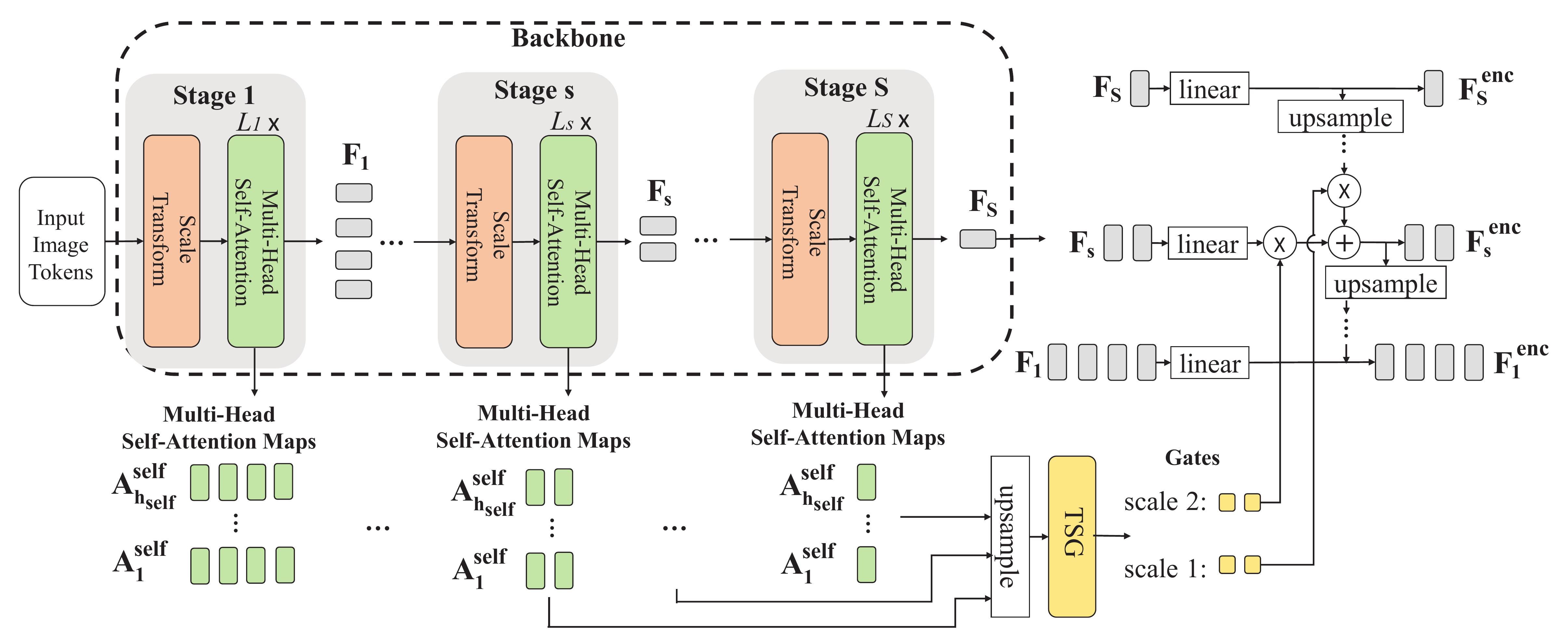}}
		\caption{Transformer Scale Gate in Encoder (TSGE). For simplicity, we only depict the self-attention module in each transformer encoder block.}
		\label{fig_method2}
	\end{figure}

	The multi-scale transformer backbone contains $S$ stages. In each stage $s$, the backbone generates a feature map $\mathbf{F_{s}} \in \mathbb{R}^{N_{s} \times d_{F,s}}$, where $d_{F,s}$ is its dimension, and $N_{s}$ is the number of patches in this feature map. Therefore, we have $S$ multi-scale features $\{\mathbf{F_{1}}, \mathbf{F_{2}},..., \mathbf{F_{S}}\}$ from the backbone model.
	We propose a TSGE module, which leverages TSG to generate scale gates $\mathbf{G}$ to refine these features, where we denote $\{\mathbf{F_{1}^{enc}}, \mathbf{F_{2}^{enc}},..., \mathbf{F_{S}^{enc}}\}$ as the refined features, $\mathbf{F_{s}^{enc}} \in \mathbb{R}^{N_{s} \times d_{F}}~(s=1,...,S)$, and all features are refined into the same dimension $d_{F}$.
	
	Inspired by feature fusion methods in CNN \cite{lin2017feature,ding2018context}, our TSGE gradually fuses small-scale features into large-scale features, as shown in Fig.~\ref{fig_method2}. Specifically, we fuse two feature maps, the smaller-scale refined feature map $\mathbf{F_{s+1}^{enc}}$ and the larger-scale feature map $\mathbf{{F}_{s}}$, at each step. 
	$\mathbf{F_{s+1}^{enc}}$ is first upsampled to fit the size of  $\mathbf{{F}_{s}}$. Then, we use a linear layer to transform the feature dimension of the larger-scale feature map $\mathbf{{F}_{s}}$. Finally, the unsampled smaller-scale feature map and the transformed larger-scale feature map are weighted by our scale gates and summed as
	\begin{equation}
		\mathbf{f_{n,s}^{enc}} = g_{n,1} \mathbf{\widetilde{f}_{n,s+1}^{enc}} + g_{n,2} \mathbf{\widetilde{f}_{n,s}}
	\end{equation}
	where $\mathbf{\widetilde{f}_{n,s+1}^{enc}} \in \mathbb{R}^{d_{F}}$ and $\mathbf{\widetilde{f}_{n,s}} \in \mathbb{R}^{d_{F}}$ are the feature vectors of the $n$-th image patch in the unsampled smaller-scale feature map and the transformed larger-scale feature map, respectively, and $\mathbf{f_{n,s}^{enc}} \in \mathbb{R}^{d_{F}}$ is the weighted sum. Weights $g_{n,1}$ and $g_{n,2}$ are generated from our TSG. The inputs of our TSG at each step are the self-attention maps corresponding to features maps we used. For example, when we fuse feature maps $\mathbf{F_{s}}$ and $\mathbf{F_{s+1}^{enc}}$, the inputs are self-attention maps in the last blocks from the $s$-th stage to the $S$-th stage, because $\mathbf{F_{s+1}^{enc}}$ has already included features from $\mathbf{F_{s+1}}$ to $\mathbf{F_{S}}$. Since these attention maps are in different sizes, we upsample them into the size of $\mathbf{{F}_{s}}$, before inputting them to TSG.
	For each image patch $n$, our TSG outputs two gates $g_{n,1}$ and $g_{n,2}$ for the two feature maps, respectively.
	
	\subsection{Transformer Scale Gate in Decoder (TSGD)}
	
	\begin{figure}[t]
		\centerline{\includegraphics[scale=0.32]{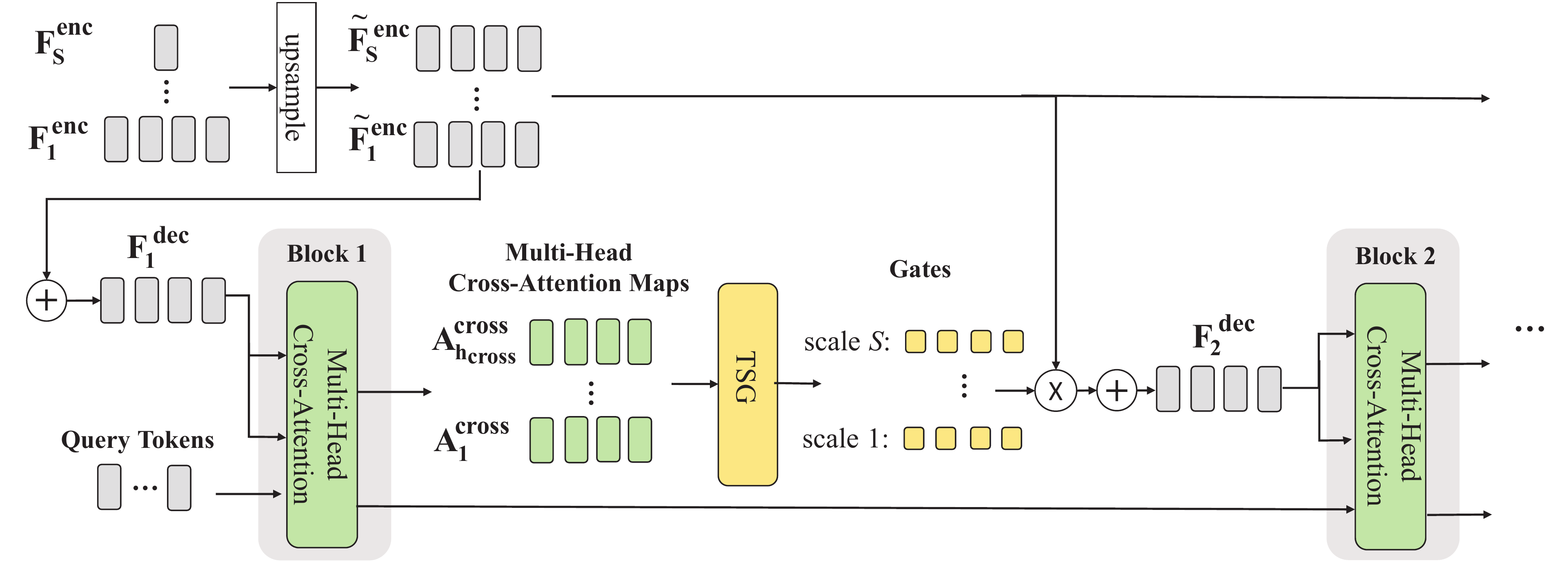}}
		\caption{Transformer Scale Gate in Decoder (TSGD). We only show the cross-attention module in each transformer encoder block for simplicity.}
		\label{fig_method3}
	\end{figure}
	
	The refined features $\{\mathbf{F_{1}^{enc}}, \mathbf{F_{2}^{enc}},..., \mathbf{F_{S}^{enc}}\}$ are then input to our decoder, and we propose a TSGD module to integrate them. In the $l$-th decoder block, to integrate multi-scale features, we first upsample them into the same size.  $\{\mathbf{\widetilde{F}_{1}^{enc}}, \mathbf{\widetilde{F}_{2}^{enc}},...,$ $\mathbf{\widetilde{F}_{S}^{enc}}\}$ are the upsampled feature maps, where $\mathbf{\widetilde{F}_{s}^{enc}} \in \mathbb{R}^{N \times d_{F}}~(s=1,...,S)$ and $N=N_1$. Then, we leverage our TSG to predict scale gates, which takes the cross-attention maps from the previous block as inputs, as shown in Fig.~\ref{fig_method3}. Our TSG outputs a matrix $\mathbf{G} \in \mathbb{R}^{N \times S}$, which contains the gates of all scales for every image patch. Finally, we uses these gates to weight the upsampled feature maps and sum them as follows:
	\begin{equation}
		\mathbf{f_{\mathit{l}, n}^{dec}} = \sum_{s=1}^{S} g_{n,s} \mathbf{\widetilde{f}_{n,s}^{enc}}
	\end{equation}
	where $\mathbf{\widetilde{f}_{n,s}^{enc}} \in \mathbb{R}^{d_{F}}$ is the feature vector of the $n$-th image patch in the feature map $\mathbf{\widetilde{F}_{s}^{enc}}$, $g_{n,s}$ in $\mathbf{G}$ represents the gate of the $s$-th scale for this patch, and $\mathbf{f_{\mathit{l}, n}^{dec}} \in \mathbb{R}^{d_{F}}$ is the weighted sum feature vector of this patch. 
	The feature map $\mathbf{F^{dec}_{l}}$ consists of $\{ \mathbf{f_{\mathit{l},1}^{dec}}, \mathbf{f_{\mathit{l},2}^{dec}}, ..., \mathbf{f_{\mathit{l},N}^{dec}} \}$, which is used to generate keys and values in the current decoder block. 
	For the first block, since there is no previous block, we only sum the upsampled multi-scale features:
	\begin{equation}
		\mathbf{f_{1, n}^{dec}} = \sum_{s=1}^{S} \mathbf{\widetilde{f}_{n,s}^{enc}}.
	\end{equation}
	
	The final segmentation result is generated by Eq.~(8), where we use the integrated feature map $\mathbf{F_{L_{dec}}^{dec}}$ in the last decoder block to generate the result:
	\begin{equation}
		\mathbf{P} = Softmax(\frac{\mathbf{F_{L_{dec}}^{dec}}\mathbf{Y}^{T}}{\sqrt{d_{F}}}) .
	\end{equation}

	\section{Experiments}
	\subsection{Experimental Settings}
	\textbf{Datasets.} We evaluate our method on two datasets, Pascal Context \cite{mottaghi2014role} and ADE20K \cite{zhou2017scene}. The Pascal Context dataset \cite{mottaghi2014role} contains 10103 images, 4998 for training and 5105 for validation. There are 60 category labels in this dataset, including 59 object classes and one background class. The ADE20K dataset \cite{zhou2017scene} includes 150 object categories, 20210 training images, 2000 validation images and 3352 testing images. For these datasets, similar to previous works, we train our method on training images and report the results on validation images.
	
	\noindent \textbf{Metrics.} We adopt the common segmentation metric, `mIoU', for evaluation, which is the average of the `IoU' values of all object classes. We report our results of a single model, without multi-scale and horizontal flip ensembles.
	
	\noindent \textbf{Implementation Details.} Our TSG can be used for any hierarchical Vision Transformer \cite{khan2021transformers}. Here, we use Swin Transformer \cite{liu2021swin} as a running example. There are four-scale feature maps in Swin Transformer \cite{liu2021swin}, i.e., $S=4$. We set the dimension $d_{F}$ of refined features to 512, use eight heads for both self- and cross-attention modules, and use three blocks in the decoder. We also set $d_{A}$ to 512, and employ a 512-dimension hidden layer in the MLP in our TSG. 
	Our model is built on the Pytorch \cite{paszke2019pytorch} platform. 
	Following common settings \cite{liu2021swin,bousselham2021efficient}, we leverage weights pretrained on ImageNet-1K and ImageNet-22K to initialize Swin-Tiny and Swin-Large, respectively. Query tokens in the decoder are initialized to zero. Other parts are randomly initialized. We adopt cross-entropy loss, `AdamW' optimizer \cite{loshchilov2018decoupled} and the `poly' learning rate decay scheduling during training, with an initial learning rate of $6 \times 10^{-5}$ and a weight decay of $10^{-2}$. 

	\subsection{Results and Comparisons}
	
	\setlength{\tabcolsep}{4pt}
	\begin{table}[t]
		\begin{center}
			\caption{Results of semantic segmentation on Pascal Context validation.}
			\label{tab_pc}
			\scalebox{0.9}{
				\begin{tabular}{L{5cm}C{2cm}C{1.5cm}C{2cm}}
					\toprule[2pt]
					Method   &Backbone &Params &mIoU(\%)  	\\
					\toprule[2pt] 
					PSPNet \cite{zhao2017pyramid} & ResNet101 & 60M & 47.0\\
					DeepLabV3+ \cite{chen2018encoder} & ResNet101 & 63M & 47.4\\
					MaskFormer \cite{cheng2021per} & ResNet101 & 60M & 53.7\\
					Segformer \cite{xie2021segformer} & MiT-B5 & 85M & 54.8\\
					SETR-MLA \cite{zheng2021rethinking} & ViT-L & 311M & 54.9 \\
					Segmenter \cite{strudel2021segmenter} & ViT-L & 333M & 58.1 \\
					Swin Transformer + UperNet \cite{liu2021swin} & Swin-Tiny & 60M & 50.2 \\
					Swin Transformer + UperNet \cite{liu2021swin} & Swin-Large & 234M & 60.3 \\
					SenFormer \cite{bousselham2021efficient} & Swin-Tiny & 144M & 53.2 \\
					SenFormer \cite{bousselham2021efficient} & Swin-Large & 314M & 62.4 \\
					\hline
					TSG (Ours)     &Swin-Tiny & 72M & 54.5 (+4.3)\\
					TSG (Ours)     &Swin-Large & 250M & \textbf{63.3} (+3.0) \\
					\bottomrule[2pt]
			\end{tabular}}
		\end{center}
	\end{table}
	\setlength{\tabcolsep}{1.4pt}
	
	\setlength{\tabcolsep}{4pt}
	\begin{table}[t]
		\begin{center}
			\caption{Results of semantic segmentation on ADE20K validation.}
			\label{tab_ade}
			\scalebox{0.9}{
				\begin{tabular}{L{5cm}C{2cm}C{1.5cm}C{2cm}}
					\toprule[2pt]
					Method   &Backbone &Params &mIoU(\%)  	\\
					\toprule[2pt]
					PSPNet \cite{zhao2017pyramid} & ResNet101 & 60M & 42.0\\
					DeepLabV3+ \cite{chen2018encoder} & ResNet101 & 63M & 45.5\\
					PVT \cite{wang2021pyramid} & PVT-Large & 65M & 42.1 \\
					Segformer \cite{xie2021segformer} & MiT-B5 & 85M & 49.6\\
					SETR-MLA \cite{zheng2021rethinking} & ViT-L & 311M & 48.6 \\
					Segmenter \cite{strudel2021segmenter} & ViT-L & 333M & 52.1 \\
					Swin Transformer + UperNet \cite{liu2021swin} & Swin-Tiny & 60M & 44.4 \\
					Swin Transformer + UperNet \cite{liu2021swin} & Swin-Large & 234M & 52.1 \\
					MaskFormer \cite{cheng2021per} & Swin-Tiny & 42M & 46.7 \\
					MaskFormer \cite{cheng2021per} & Swin-Large & 212M & 54.1 \\
					SenFormer \cite{bousselham2021efficient} & Swin-Tiny & 144M & 46.0 \\
					SenFormer \cite{bousselham2021efficient} & Swin-Large & 314M& 53.1 \\
					\hline
					TSG (Ours)     &Swin-Tiny & 72M & 47.5 (+3.1)\\
					TSG (Ours)     &Swin-Large & 250M  & \textbf{54.2} (+2.1)\\
					\bottomrule[2pt]
			\end{tabular}}
		\end{center}
	\end{table}
	\setlength{\tabcolsep}{1.4pt}
	
	Table~\ref{tab_pc} shows the results of existing state-of-the-art methods and our method on the Pascal Context dataset. Compared with our baselines, Swin Transformer \cite{liu2021swin} Tiny and Large, our method achieves 4.3\% and 3.0\% gains, respectively. Compared with SenFormer \cite{bousselham2021efficient}, which obtains the best performance among existing approaches, our proposed method yields improvements of 1.3\% and 0.9\% when using Swin-Tiny and Swin-Large backbones, respectively. However, the number of parameters of our method is significantly lower than that of SenFormer \cite{bousselham2021efficient}. SenFormer \cite{bousselham2021efficient} uses a heavyweight architecture with multiple transformer decoders to generate multi-scale predictions. In contrast, our method only leverages one decoder and lightweight scale gates, while achieving better performance.
	
	We report the results on the ADE20K dataset in Table~\ref{tab_ade}. It can be seen that our modules achieve consistent gains over baselines,  i.e., Swin Transformer \cite{liu2021swin} Tiny and Large, by 3.1\% and 2.1\%, respectively. Our method also outperforms the best prior work, MaskFormer \cite{cheng2021per}, by 0.8\% (Swin-Tiny backbone) and 0.1\% (Swin-Large backbone). MaskFormer \cite{cheng2021per} designs a powerful decoder, while we aim to propose a general purpose multi-scale feature selection module which can be used in a plug-and-play manner to enhance existing transformer segmentation architectures. Our module can be combined with MaskFormer \cite{cheng2021per}.
	
	\begin{figure}[t]
		\centerline{\includegraphics[scale=0.32]{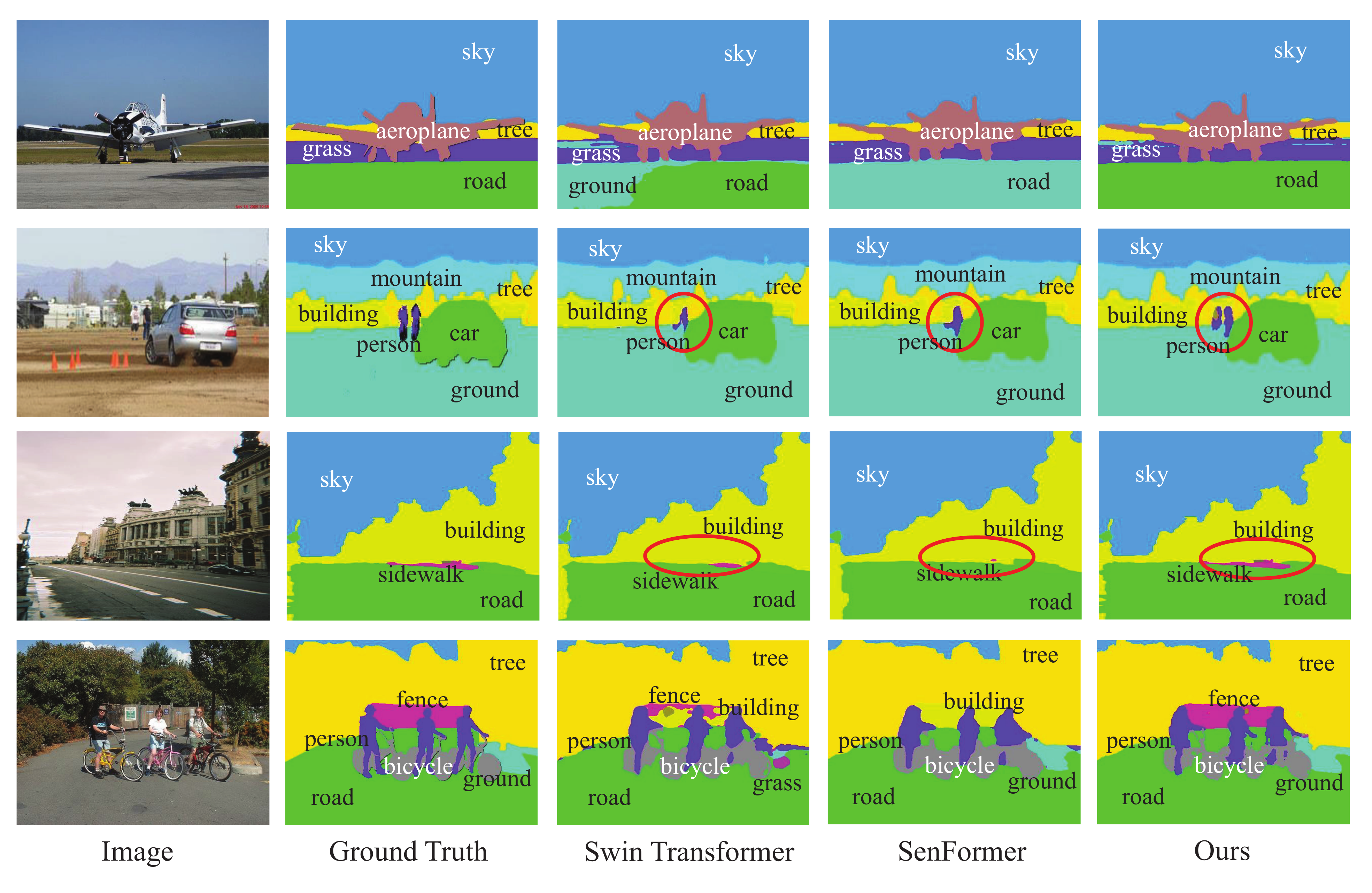}}
		\caption{Qualitative results on Pascal Context validation. Left to right: input images, ground truths, the results from Swin Transformer \cite{liu2021swin}, SenFormer \cite{bousselham2021efficient} and our TSG. All methods use Swin-Large as the backbone.}
		\label{fig_results}
	\end{figure}
	
	Fig.~\ref{fig_results} depicts qualitative results on the Pascal Context dataset. Previous methods such as Swin Transformer \cite{liu2021swin} and SenFormer \cite{bousselham2021efficient} only simply combine multi-scale features or multi-scale segmentation results without any selection, and thus fail to segment many small objects, such as the left `person' in the second image and the `sidewalk' in the third image in Fig.~\ref{fig_results}. Our TSG is able to select suitable scales for image patches. When segmenting patches including small objects, our approach selects high-resolution features based on transformer attention cues. Therefore, our method successfully segments these small objects. Previous approaches are also prone to over-segmentation and mis-recognitions. For example, in the first image in Fig.~\ref{fig_results}, Swin Transformer \cite{liu2021swin} over-segments the `road' object as two objects, and SenFormer \cite{bousselham2021efficient} mis-recognizes the `road' object as `ground'. Similarly, in the fourth image in Fig.~\ref{fig_results}, Swin Transformer \cite{liu2021swin} and SenFormer \cite{bousselham2021efficient} also fail to segment the `fence' object. Our method avoids these, by selecting suitable scales to segment and recognize objects of diverse scales.
	
	\subsection{Ablation Analysis}
	
	\setlength{\tabcolsep}{4pt}
	\begin{table}[t]
		\begin{center}
			\caption{The effects of main components in our method on Pascal Context validation.}
			\label{tab_TSGE}
			\scalebox{0.95}{
				\begin{tabular}{C{1.5cm}L{3cm}L{4.2cm}C{2cm}}
					\toprule[2pt]
					Model & Encoder & Decoder &mIoU(\%)  	\\
					\toprule[2pt]
					1& Swin-Tiny & Upernet  & 50.2 \\
					2& Swin-Tiny + TSGE & Upernet   & 51.6 (+1.4) \\
					\hline
					3& Swin-Tiny  &  Baseline Decoder  & 51.3 \\
					4& Swin-Tiny  + TSG  & Baseline Decoder  & 52.4 (+1.1) \\
					5& Swin-Tiny & Baseline Decoder + TSGD & 52.7 (+1.4)\\
					\hline
					6& Swin-Tiny + FPN   & Baseline Decoder  & 52.9 \\
					7& Swin-Tiny + TSGE & Baseline Decoder  & 53.9 (+1.0)\\
					8& Swin-Tiny + FPN & Baseline Decoder + TSGD & 54.1 (+1.2) \\
					\hline
					9& Swin-Tiny + TSGE  & Baseline Decoder + TSGD & 54.5 (+4.3) \\
					\bottomrule[2pt]
			\end{tabular}}
		\end{center}
	\end{table}
	\setlength{\tabcolsep}{1.4pt}
	
		\setlength{\tabcolsep}{4pt}
	\begin{table}[t]
		\begin{center}
			\caption{The effects of features on different scales.}
			\label{tab_scale}
			\scalebox{0.82}{
				\begin{tabular}{lccccc}
					\toprule[2pt]
					Model& $F_{1}$ (1/4) & $F_{2}$ (1/8) & $F_{3}$ (1/16) & $F_{4}$ (1/32) &mIoU(\%)  	\\
					\toprule[2pt]
					Swin-Tiny + FPN + Baseline Decoder && &  & \checkmark & 47.2 \\ 
					Swin-Tiny + FPN + Baseline Decoder && & \checkmark & & 50.1 \\
					Swin-Tiny + FPN + Baseline Decoder && \checkmark & & & 51.6 \\
					Swin-Tiny + FPN + Baseline Decoder &\checkmark & & & & 52.5\\
					Swin-Tiny + FPN + Baseline Decoder & \checkmark & \checkmark & \checkmark & \checkmark & 52.9\\
					Swin-Tiny + TSGE + Baseline Decoder + TSGD & \checkmark & \checkmark & \checkmark & \checkmark & 54.5\\
					\bottomrule[2pt]
			\end{tabular}}
		\end{center}
	\end{table}
	\setlength{\tabcolsep}{1.4pt}
	
	\setlength{\tabcolsep}{4pt}
	\begin{table}[t]
		\begin{center}
			\caption{The effects of different designs in our TSG on Pascal Context validation.}
			\label{tab_settings}
			\scalebox{0.95}{
				\begin{tabular}{C{7cm}C{2cm}}
					\toprule[2pt]
					Design &mIoU(\%)	\\
					\toprule[2pt]
					TSG with multi-head average  & 54.1  \\ 
					TSG with multi-head concatenation & 54.5 (+0.4) \\
					\hline
					TSGs with shared weights & 53.8 \\
					TSGs with independent weights & 54.5 (+0.7)\\
					\bottomrule[2pt]
			\end{tabular}}
		\end{center}
	\end{table}
	\setlength{\tabcolsep}{1.4pt}
	
	\begin{figure}[t]
		\centerline{\includegraphics[scale=0.32]{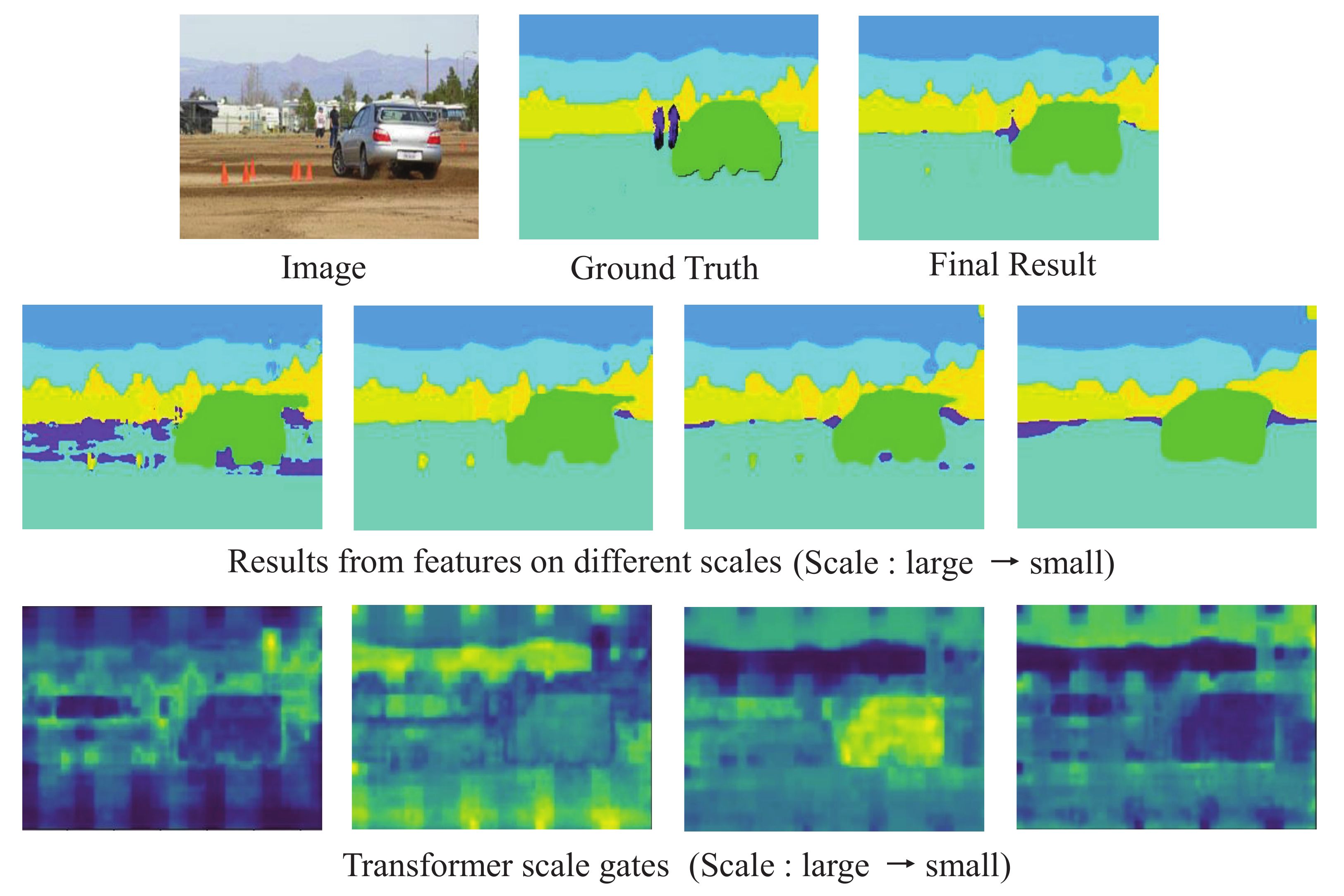}}
		\caption{Visualization of our scale gates and the results from features on different scales. All methods take Swin-Tiny as the backbone.}
		\label{fig_ablation}
	\end{figure}

	\textbf{\underline{TSG in encoder.}} We conduct multiple ablations to verify the contributions of our TSG module for the Transformer encoder in Table~\ref{tab_TSGE}. The vanilla Swin Transformer \cite{liu2021swin} (Model 1) takes UperNet \cite{xiao2018unified} as its decoder, which already includes FPN \cite{lin2017feature} and PPM \cite{zhao2017pyramid}. In Model 2, we use our TSGE to replace FPN and PPM in UperNet. From the first and second lines in Table~\ref{tab_TSGE}, it can be observed that our method obtains an improvement of 1.4\% in this setting. We next test models using our baseline decoder. Our baseline decoder requires to first integrate multi-scale feature maps into one feature map. Thus, we first add a linear layer to every feature map to convert their dimensions into the same. Then, the converted multi-scale feature maps are upsampled into the same size, and summed as the input of our baseline decoder. In Model 4, our TSG generates scale gates from self-attention modules in the encoder, and these gates are used to weight the multi-scale feature maps before summing them. Through our TSG, the mIoU can be improved by 1.1\%. In Models 6\&7, we use FPN \cite{lin2017feature} and our TSG to refine multi-scale feature maps, respectively, and sum the refined feature maps in our decoder. Compared with FPN, our TSGE achieves a gain of 1.0\%. 
	
	\noindent \textbf{\underline{TSG in decoder.}} From Models 3\&5 in Table~\ref{tab_TSGE}, we observe that our TSGD outperforms the baseline decoder by 1.4\% when using the `Swin-Tiny' encoder. With the `Swin-Tiny + FPN' encoder (Models 6\&8), our TSGD improves the performance by 1.2\%. Our final model (Model 9) uses both TSGE and TSGD, which achieves an improvement of 4.3\%, compared with Swin Transformer \cite{liu2021swin}. 
	These results suggest that our proposed TSG is effective in fusing cues across multiple spatial resolutions. 
	
	\noindent \textbf{\underline{Results on different scales.}} We compare the results from multi-scale features in Table~\ref{tab_scale}. Our method significantly outperforms all singe-scale models and the approaches that simply combine the multi-scale features, benefiting from our transformer-based scale selection.
	
	\noindent \textbf{\underline{Dissecting TSG.}} Table~\ref{tab_settings} shows the results of our TSG with different settings. In Sec.~3.2, we concatenate multi-head attention maps, which improves the mIoU by 0.4\%, compared with averaging multi-head attention maps. This is because some information may be lost in the average, while the concatenation retains all information in attention maps. We use multiple TSG with independent weights in different steps in the encoder and different blocks in the decoder. Compared with sharing weights among different TSGs, independent weights achieve 0.7\% improvements, because every TSG has different inputs and independent weights show better ability for different inputs. Nonetheless, our TSG with shared weights can also improve the performance by 3.6\%, compared with Swin Transformer \cite{liu2021swin} baseline.
	
	\noindent \textbf{\underline{Transformer scale gates.}} We visualize the results from multi-scale features and our scale gates in the second decoder block in Fig.~\ref{fig_ablation}. From the 2nd row (Fig.~\ref{fig_ablation}), we find that large-scale features generate fine-grained segmentation results, but they cause over-segmentations.
	Although small-scale features can avoid over-segmentations, they are unable to capture object boundaries and small objects.
	From the 3rd row, we observe that our TSG highlights object boundaries and small objects in large-scale feature maps to obtain fine-grained segmentations, while selecting small-scale features for large objects to reduce over-segmentations.
	
	\begin{figure}[t]
		\centerline{\includegraphics[scale=0.28]{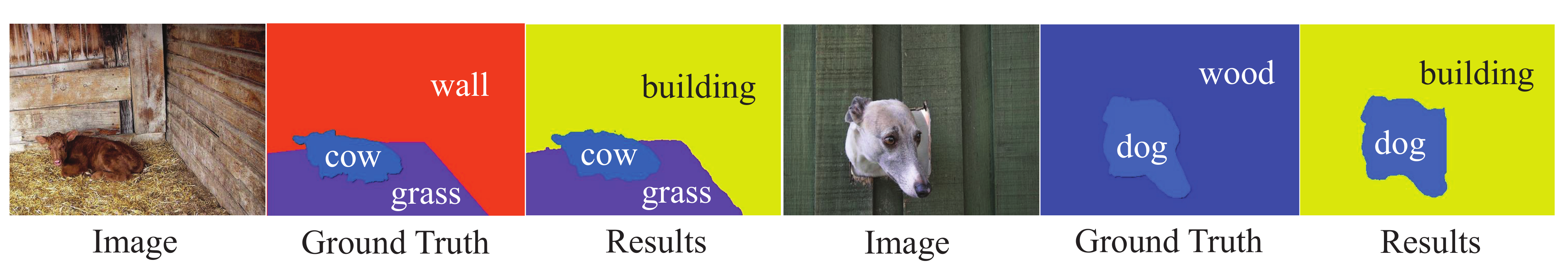}}
		\caption{Failure cases from our method with the Swin-Large backbone.}
		\label{fig_failure}
	\end{figure}
	
	\noindent \textbf{\underline{Failure cases.}} Fig.~\ref{fig_failure} shows some failure cases of our method. Although our TSG is able to reduce over-segmentations, under-segmentations and mis-recognitions caused by sub-optimal scales, TSG still confuses amongst some objects with similar appearances, e.g., `wood', wooden `wall' and `building'. These confusions can be alleviated by using more powerful backbones with discriminative features.
	
	\section{Conclusion}
	In this paper, we have presented a Transformer Scale Gate (TSG) module, which exploits inherent properties in Vision Transformers to effectively select multi-scale features for semantic segmentation.
	Our TSG is a simple transformer-based module with lightweight linear layers, which can be used in transformer segmentation networks in a plug-and-play manner.
	We have also proposed TSGE and TSGD, which leverage our TSG to further improve the segmentation accuracy in the transformer encoder and decoder, respectively.
	TSGE refines multi-scale features in the encoder by the self-attention guidance, while TSGD integrates multi-scale features in the decoder based on cross-attention maps.
	Extensive experiments on two semantic segmentation datasets demonstrate the effectiveness of our proposed method.

	\clearpage
	%
	%
	\bibliographystyle{splncs04}
	\bibliography{my_reference}
\end{document}